\documentclass[letterpaper, 10 pt, conference]{ieeeconf} 

\IEEEoverridecommandlockouts                              

\usepackage{graphicx} 
\usepackage[caption=false,font=footnotesize]{subfig}
\usepackage{mathptmx} 
\usepackage{mathrsfs}
\usepackage{mathtools}
\usepackage{amsmath} 
\usepackage{amssymb} 
\usepackage{tabularx}
\usepackage[ruled,linesnumbered,noend]{algorithm2e}
\usepackage{pgfplots}
\usepackage{cite}
\pgfplotsset{compat=1.13}

\DeclareMathAlphabet{\mathcal}{OMS}{cmsy}{m}{n}

\title{\LARGE \bf
Planning Optimal Trajectories for Mobile Manipulators under 
End-effector Trajectory Continuity Constraint
}

\author{Quang-Nam Nguyen$^{1}$ and Quang-Cuong Pham$^{1,2}$%
\thanks{$^{1}$Singapore Centre for 3D Printing (SC3DP), Nanyang Technological 
        University (NTU), Singapore {\tt\small nam.ngquang@gmail.com}}%
\thanks{$^{2}$Eureka Robotics, Singapore}%
}

\newtheorem{definition}{Definition}
\newtheorem{theorem}{Theorem}

\begin{document}
\bstctlcite{BSTcontrol}

\maketitle
\thispagestyle{empty}
\pagestyle{empty}

\begin{abstract}
        Mobile manipulators have been employed in many applications that are 
        traditionally performed by either multiple fixed-base robots or a large 
        robotic system. 
        This capability is enabled by the mobility of the mobile base. 
        However, the mobile base also brings redundancy to the system, 
        which makes mobile manipulator motion planning more challenging. 
        In this paper, we tackle the mobile manipulator motion planning problem 
        under the end-effector trajectory continuity constraint in which the 
        end-effector is required to traverse a continuous task-space trajectory 
        (time-parametrized path), such as in mobile printing or spraying applications. 
        Our method decouples the problem into: 
        (1) planning an optimal base trajectory subject to geometric task 
        constraints, end-effector trajectory continuity constraint, collision 
        avoidance, and base velocity constraint; which ensures that 
        (2) a manipulator trajectory is computed subsequently based on the 
        obtained base trajectory. 
        To validate our method, we propose a discrete optimal base trajectory 
        planning algorithm to solve several mobile printing tasks in hardware 
        experiment and simulations.
\end{abstract}

\section{Introduction}

A mobile manipulator often consists of one (or multiple) manipulator, such as a 
robotic arm, mounted on a mobile base which helps extend the workspace of the 
manipulator. 
This mobility enables mobile manipulators to be employed in many large-scale 
applications which are usually performed by multiple robots or a larger system, 
such as in printing \cite{sustarevas2022autonomous,sustarevas2021task,
tiryaki2019printing,zhang2018large,dorfler2022additive}, pick and place 
\cite{malhan2022finding,xu2021planning}, drilling \cite{nguyen2023task}, etc. 
However, the mobile base adds redundant degrees of freedom (DOFs) to the system, 
which makes the dimension of the configuration space higher than that of the 
task space. 
On one hand, this redundancy brings challenges to task and motion planning, 
but on the other hand, it leaves room for optimization \cite{lynch2017modern}.

In general, a robot accomplishes a high-level task (e.g. create multiple holes 
on a workpiece) by performing one or a sequence of sub-tasks (e.g. drill a hole, 
move to the next target). 
The high-level task is solved by finding the feasible or optimal sequence to 
perform the sub-tasks, which can be planned separately (task sequencing before 
motion planning) or in combined task and motion planning \cite{srivastava2014combined}.

\begin{figure}[tb]
        \centering
        \includegraphics[width=1.\linewidth]{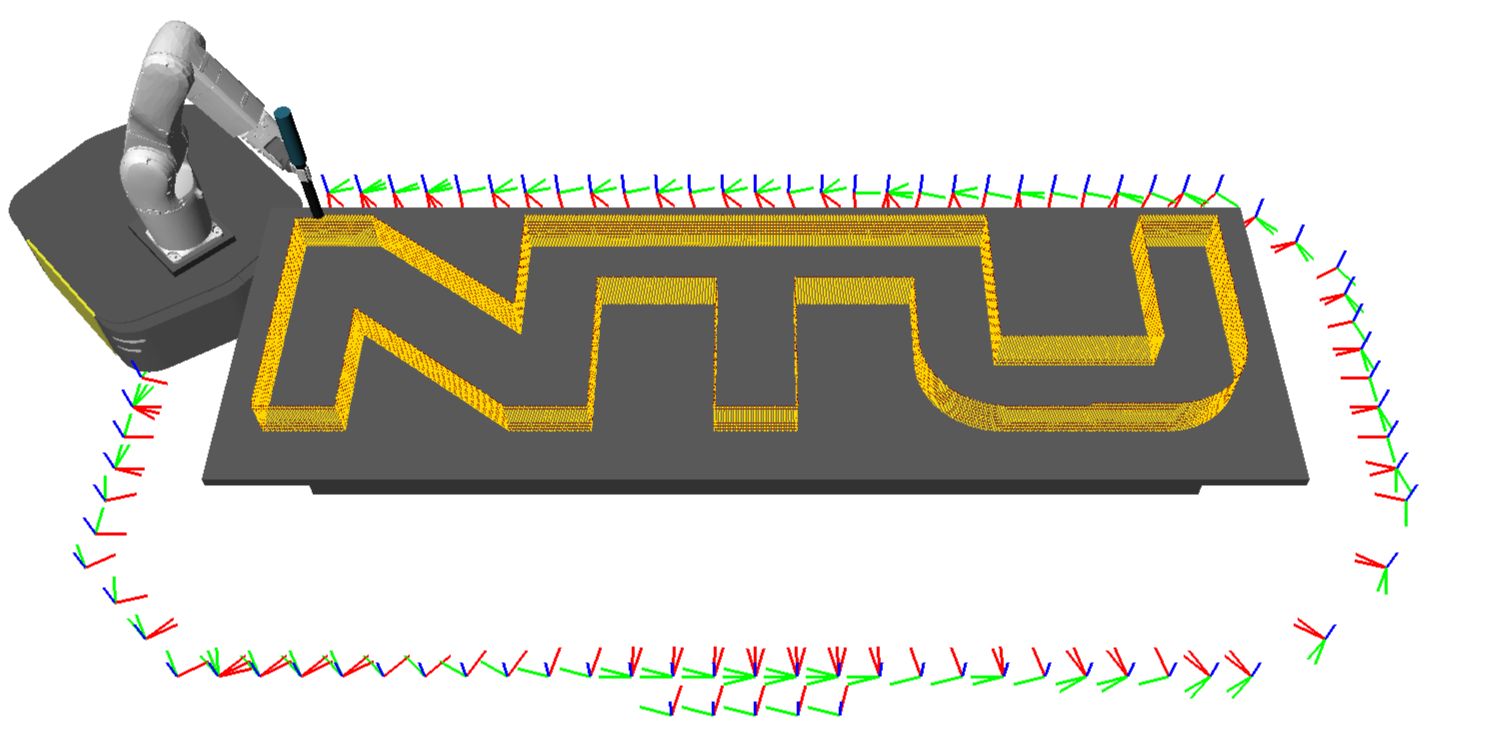}
        \caption{Mobile 3D printing of NTU shape (10 layers, size: 
        $3 \times 0.75 \times 0.15m$, total printing path length: 112.9m) 
        with constant nozzle speed of $10cm/s$.}
        \label{fig:3Dprinting_NTU}
\end{figure}

Due to the localization uncertainty of the mobile base (typically centimetres), 
it is often desirable to optimize the base motion based on an objective such as 
travel distance, control effort, number of base stops, etc. 
Optimization can be done in task sequencing and/or motion planning 
\cite{lavalle2006planning}. 
Optimal mobile manipulator task sequencing has been addressed in 
many applications such as pick and place, drilling, scanning 
\cite{nguyen2023task,xu2021planning, malhan2022finding}. 
For optimization in motion planning, an example is mobile printing 
(Fig. \ref{fig:3Dprinting_NTU}) in which the main task is to move 
the nozzle along a printing trajectory. 
In this case, we desire a slow and stable base motion while printing.

In this paper, we focus on motion planning for motions that involve the 
\textit{end-effector trajectory continuity constraint} which requires the 
end-effector to traverse a continuous trajectory (time-parametrized path) 
such as in mobile printing \cite{tiryaki2019printing,sustarevas2021task, 
dorfler2022additive,sustarevas2022autonomous} or spraying \cite{jenny2023continuous}.
Note that this end-effector trajectory-continuous motion is different from the 
end-effector path-continuous motion in which the end-effector is only required 
to traverse a continuous path without specified velocity or timing information 
\cite{nagatani2002motion,oriolo2002probabilistic,oriolo2005motion, 
welschehold2018coupling,haviland2022holistic}. 
There are two approaches in mobile manipulator motion planning: whole-body 
planning and decoupled planning. 
These differences will be discussed in the next section.

Our contribution is an optimal base trajectory planning method for mobile 
manipulator motion planning under end-effector trajectory continuity constraint, 
which consists in:
\begin{itemize}
        \item Using the manipulator's reachability to confine the base configurations 
        inside an admissible region in configuration spacetime that satisfies 
        end-effector trajectory continuity constraint, geometric task constraints, 
        and collision avoidance; which ensures a manipulator trajectory can be 
        planned subsequently.
        \item Proposing a discrete optimal base trajectory planning algorithm to 
        solve the above problem with considerations for the base velocity constraint 
        and optimization.
\end{itemize}

The remainder of this paper is organized as follows. 
In section II, we discuss related works.
Section III formulates the problem and gives an overview of our method. 
In section IV, we propose a discrete optimal base trajectory planning algorithm. 
Section V validates our method in several mobile printing tasks and provides some 
evaluations.

\section{Related Works}

A recent survey on mobile manipulator motion planning can be found in 
\cite{sandakalum2022motion}. 
In some applications, short-term whole-body planning can be performed online during 
control \cite{avanzini2015constraint,pankert2020perceptive,giftthaler2017efficient}. 
In other applications that require long-term plans, high dimensionality becomes 
a challenge even for offline planning. 
To tackle this, an effective approach is to decouple the whole-body problem into 
base's and manipulator's motion planning problems which are, however, not entirely 
separated since their trajectories must ensure manipulator's reachability 
(under joint limits). 
This relationship can be represented by reachability map 
\cite{zacharias2008positioning}, inverse reachability map \cite{vahrenkamp2013robot}, 
or other geometric approximations \cite{nguyen2023task,welschehold2018coupling}. 

In our work, we use the geometric reachable region method \cite{nguyen2023task} 
which is approximately equivalent to reachability and inverse reachability maps 
but has some advantages. 
Firstly, representing reachability by a geometric shape with geometric parameters 
(see Fig. \ref{fig:kr}) is light-weight compared to storing a reachability or 
inverse reachability map and matching them with the floor. 
Secondly, Inverse Kinematics (IK) solutions exist everywhere inside the 
geometric reachable region with 100\% certainty, although in return for that, 
the end-effector's orientation must be within a specified range. 
A limitation of all above-mentioned methods is that they do not guarantee a 
continuous joint trajectory between IK solutions \cite{xian2017closed}, 
which needs further research in the future.

In end-effector path-continuous tasks, the end-effector is required to track a 
given path. 
Using the decoupled approach, the base motion planning can be treated as a path 
planning problem and solved using variations of Rapidly-exploring Random Tree 
(RRT) planners \cite{oriolo2002probabilistic,oriolo2005motion}. 
In the whole-body approach, the whole-body motion can be short-term planned and 
controlled online using Quadratic Programming (QP) \cite{haviland2022holistic}, 
Model Predictive Control (MPC) \cite{avanzini2015constraint,pankert2020perceptive}, 
or Sequential Linear Quadratic Optimal Control (SLQ) \cite{giftthaler2017efficient}.

For end-effector trajectory-continuous tasks, the difference from path-continuous 
tasks is the time-parametrization along end-effector's trajectory. 
A major application of this class of tasks is mobile printing/spraying. 
Initial works in this field were performed with the base trajectory being 
planned manually \cite{tiryaki2019printing,jenny2023continuous}. 
More recently in \cite{sustarevas2021task,sustarevas2022autonomous}, the authors 
followed the decoupled approach and used a two-step base motion planning method. 
Their method consists of a base path planning step using a variation of RRT* planner 
and a post-processing step that smooths and time-parametrizes the obtained path. 
However, the base kinematic constraints (e.g. velocity limits) were not considered 
during path planning which may results in unsuccessful post-processing. 
On the other hand, in the whole-body approach, while there are methods to compute 
pathwise-IK solutions for redundant systems \cite{rakita2019stampede,rakita2018relaxedik}, 
extending these methods to mobile manipulation is not trivial since it will also 
face the challenge in handling kinematic constraints and optimization for the base. 
This direction can be investigated in future works.

In this paper, we would like to improve the decouple approach to address the 
above-mentioned limitations by formulating the base motion planning problem 
as a constrained optimal trajectory planning problem which allows us to consider 
optimization and multiple constraints (end-effector trajectory continuity constraint, 
geometric task constraints, collision avoidance, and base velocity constraint).

\section{Methodology Overview}

\subsection{Configuration spacetime}

The \textit{configuration space (C-space)} $\mathcal{C}$ of a system is an 
$n$-dimensional manifold in which each point has $n$ \textit{generalized coordinates} 
representing the system configuration:
\begin{equation} \label{eq:cspace}
        \mathbf{q} \coloneqq (q_1, ..., q_n) \in \mathcal{C}
\end{equation}

Spacetime is a representation of the evolution of a dynamical system through time.
The term \textit{configuration spacetime (C-spacetime)} was first used in 1939 
\cite{trumper1983lagrangian} and has been defined as the topological product of 
a real time axis $\mathcal{T} \subseteq \mathbb{R}$ and the configuration space: 
$\mathcal{X} \coloneqq \mathcal{T} \otimes \mathcal{C}$.
Thus, the configuration spacetime is an $(n+1)$-manifold:
\begin{equation}
        \mathcal{X} = \{ \mathbf{x} \coloneqq (t, \mathbf{q}) \;|\; 
        t \in \mathcal{T}, \mathbf{q} \in \mathcal{C} \}
\end{equation}
where each point $\mathbf{x} = (t, q_1, ..., q_n)$ is called an \textit{event} 
which occurs at time $t$ (\textit{temporal coordinate}) when the system is 
having the configuration $q_1, ..., q_n$ (\textit{spatial coordinates}).

\begin{figure}[tb]
        \centering
        \includegraphics[width=0.85\linewidth]{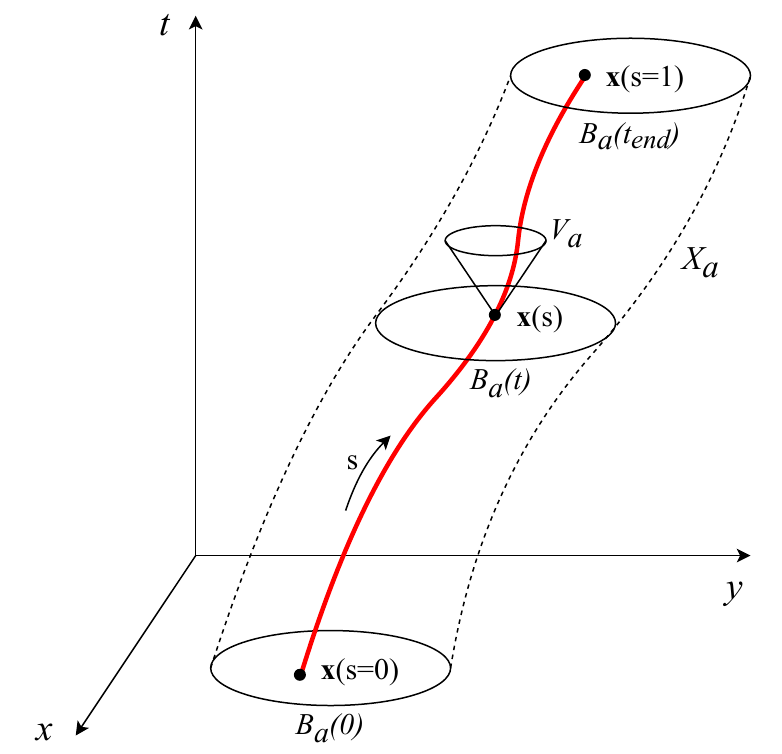}
        \caption{Visualization of a trajectory in the base configuration 
        spacetime (B-spacetime). 
        The base orientation $\varphi$ axis is not shown. 
        Any point $\mathbf{x}(s)$ in the trajectory must be inside an 
        admissible B-space $\mathcal{B}_a(t)$ and its velocity $\dot{\mathbf{x}}(s)$ must be 
        inside a cone of admissible spacetime velocities $\mathcal{V}_a$.
        The admissible B-spacetime $\mathcal{X}_a$ is obtained as the 
        admissible B-space changes in time.}
        \label{fig:Bspacetime}
\end{figure}

The \textit{trajectory} of a system is a continuous sequence of events, 
which is a curve $\mathbf{q}(t)$, or equivalently, $\mathbf{x}(s)$ in C-spacetime. 
Here, the path parameter $s$ goes from $s=0$ at a start event to $s=1$ at an 
end event (see Fig. \ref{fig:Bspacetime}).
The tangent vector at any point along the trajectory:
\begin{equation}
        \mathbf{x}' \coloneqq \frac{d \mathbf{x}}{ds} =
        \begin{pmatrix}
                t'\\
                \mathbf{q}'
        \end{pmatrix}
\end{equation}
must satisfy $t' \coloneqq dt/ds > 0$ since time is monotonic.

The \textit{spacetime velocity} at any point along the trajectory is parallel 
to the tangent vector at that point:
\begin{equation}
        \dot{\mathbf{x}} \coloneqq \frac{d \mathbf{x}}{dt} = \frac{\mathbf{x}'}{t'} = 
        \begin{pmatrix}
                1\\
                \dot{\mathbf{q}}
        \end{pmatrix}
\end{equation}
where $\dot{\mathbf{q}} \coloneqq d\mathbf{q}/dt$ is called the 
\textit{generalized velocity}.

\subsection{Mobile manipulators in configuration spacetime}

For example, our mobile manipulator consists of a $6$-DOF manipulator 
$\mathbf{q}_m = (\theta_1, ..., \theta_6)$ and a 3-DOF planar mobile base 
$\mathbf{q}_b = (x, y, \varphi)$ where $\varphi \in \mathbb{S}^1$ is the 
orientation (yaw angle) of the base.
Therefore, the whole-body configuration and C-space of a mobile manipulator are: 
\begin{equation}
        \mathbf{q} = (\theta_1, ..., \theta_6, x, y, \varphi) \in \mathcal{C}, 
        \quad \mathcal{C} \subseteq \mathbb{R}^8 \otimes \mathbb{S}^1
\end{equation}

We follow the decoupled approach which treats the manipulator and the mobile 
base separately, i.e. $\mathcal{C} = \mathcal{M} \otimes \mathcal{B}$ where the 
\textit{manipulator configuration space} is $\mathcal{M} \subseteq \mathbb{R}^6$ 
and the \textit{base configuration space (B-space)} is 
$\mathcal{B} \subseteq \mathbb{R}^2 \otimes \mathbb{S}^1$.

We define the \textit{base configuration spacetime (B-spacetime)}:
\begin{equation}
        \mathbf{x} = (t, x, y, \varphi) \in \mathcal{X}, \quad
        \mathcal{X} \subseteq \mathbb{R}^3 \otimes \mathbb{S}^1
\end{equation}
where we denote $\mathbf{x}, \mathcal{X}$ instead of $\mathbf{x}_b, \mathcal{X}_b$
since we only need to apply the concept of spacetime on the mobile base.

\subsection{Constraints}

\subsubsection{Geometric task constraints}
In a 6D task, the end-effector follows as rigid body motions, so the task 
space is generally $SE(3)$. 
For 5D tasks, the geometric task constraints are end-effector's 3D position and 
vector orientation, so the task space is $\mathbb{R}^3 \otimes \mathbb{S}^2$. 
For example, printing is a 5D task in which the rotation around nozzle's axis is free. 
In this paper, we consider 5D tasks which bring more redundant DOFs to the problem 
($n-5$ compared to $n-6$). 
Moreover, we notice that in printing, the nozzle's axis usually points vertically 
downwards which is parallel to the base's axis or only tilts slightly from it. 
This makes mobile printing tasks more challenging because the base can orient 
towards each end-effector pose at any angle: $\varphi \in \mathbb{S}^1$.

\subsubsection{Collision avoidance}
The base configuration spacetime allows us to consider collisions between the base 
and static or known dynamic obstacles by setting a safe distance from them. 
From static obstacles, we get an initial set of collision-free base configurations: 
$\mathcal{B}_{free}(0) \subseteq \mathcal{B}$. 
From known dynamic obstacles (e.g. printed structure), the sets of collision-free 
base configurations are obtained at different time instances: 
\begin{equation}
        \mathcal{B}_{free}(t) \subseteq \mathcal{B}_{free}(0), \; t \in \mathcal{T}
\end{equation}

\subsubsection{End-effector trajectory continuity constraint}
The end-effector is required to follow a continuous task-space trajectory 
$\mathbf{p}(t)$ instead of just a geometric path. 
For each end-effector pose in its task-space trajectory $\mathbf{p}(t)$, the 
base configuration $\mathbf{q}_b(t)$ must be kept within a set of admissible 
configurations $\mathcal{B}_a(t) \subseteq \mathcal{B}_{free}(t)$ so that the 
end-effector can reach the pose without violating joint limits.
The set of admissible events is therefore a subset of the B-spacetime which we 
call the \textit{admissible B-spacetime}: (see Fig. \ref{fig:Bspacetime})
\begin{equation}
        \mathcal{X}_a = \{ \mathbf{x} = (t,\mathbf{q}_b) \;|\; 
        t \in \mathcal{T},\; \mathbf{q}_b \in \mathcal{B}_a(t) \} 
        \subseteq \mathcal{X}
\end{equation}
The admissible B-spacetime will be obtained based on the reachability analysis 
which will be explained in section VI.

\subsubsection{Base velocity constraint}
We consider the translational and rotational velocity limits of the mobile base 
which can be represented by a \textit{set of admissible spacetime velocities}:
\begin{equation} \label{eq:admiss_vel}
        \mathcal{V}_a = \bigl\{ \dot{\mathbf{x}} \coloneqq (1, \dot{\mathbf{q}}_b) \;|\; 
        \dot{x}^2 + \dot{y}^2 \leq v_{max}^2,\;
        \dot{\varphi}^2 \leq \omega_{max}^2 \bigr\}
\end{equation}
This constraint is visualized in Fig. \ref{fig:Bspacetime}: at every point 
on the trajectory, the spacetime velocity must be kept inside a cone.

\subsection{Problem formulation}

The goal of mobile manipulator motion planning is to find the whole-body joint 
trajectory so that the robot traverses the end-effector task-space trajectory 
subject to constraints and optimization. 
We decouple this problem into manipulator's and base's trajectory planning problems 
as follows.

The base trajectory planning step is solved first, during which we use the manipulator's 
kinematic reachability and other constraints to confine the base configurations 
inside an admissible B-spacetime so that the manipulator can reach the desired pose 
at any given time without violating joint limits.

\subsubsection{Base trajectory planning}
The \textit{feasible trajectory planning} problem for mobile base is to find 
a base trajectory inside the admissible B-spacetime, such that the velocity 
at every trajectory point is an admissible velocity:
\begin{equation}
        \mathbf{x}(s) \in \mathcal{X}_a, \quad
        \dot{\mathbf{x}}(s) \in \mathcal{V}_a \quad
        \forall s \in [0,1]
\end{equation}

Handling constraints: geometric task constraints, end-effector trajectory 
continuity constraint, and collision avoidance are considered in $\mathcal{X}_a$; 
while base velocity constraint is realized by $\mathcal{V}_a$.

\textit{Optimal trajectory planning}: among feasible solutions, it is desirable 
to find an optimal trajectory to minimize a cost:
\begin{equation}
        \mathbf{x}^{opt}(s) = \arg\min_{\mathbf{x}(s)} J[\mathbf{x}(s)]
\end{equation}
where $J[\mathbf{x}(s)]$ is the cost functional, typically formulated as an 
integral of a Lagrangian.

We use the cost functional for minimum control effort which is defined as:
(we denote $a \cdot B \cdot a \coloneqq a^T B a$)
\begin{equation}
\begin{aligned} \label{eq:cost}
        J[\mathbf{q}_b(t)] \coloneqq & \int_{0}^{t_{end}} 
        L(\mathbf{q}_b, \dot{\mathbf{q}}_b, t) dt 
        = \int_{0}^{t_{end}} 
        \dot{\mathbf{q}}_b \cdot \mathbf{I}_q \cdot \dot{\mathbf{q}}_b \,dt \\
        \Leftrightarrow \quad
        J[\mathbf{x}(s)] = & \int_{0}^{1} 
        \mathbf{x}' \cdot \mathbf{I}_x \cdot \mathbf{x}' \,\frac{ds}{t'}, \quad
        \mathbf{I}_x =
        \begin{pmatrix}
                0 & 0 \\
                0 & \mathbf{I}_q
        \end{pmatrix}
\end{aligned}
\end{equation}
where the weight matrix $\mathbf{I}_q \coloneqq diag(1,1,w)$ sets the relative 
weight between rotational and translational base motions.
This cost functional for minimum control effort effectively prevents unnecessarily 
large velocity changes, since a slowe and stable velocity profile results in a 
low control effort.

\subsubsection{Manipulator trajectory planning}
Given the obtained base trajectory, the manipulator trajectory planning problem 
is to find the joint trajectory of the manipulator so that its end-effector follows 
the desired task-space trajectory.

The method of using manipulator's reachability to confine the admissible base 
configurations guarantees that IK solutions exist for for every pair of base 
and end-effector poses $\mathbf{q}_b(t), \mathbf{p}(t)$. 
Thus, the joint trajectory can be computed using fast IK solvers such as 
OpenRAVE's IKFast \cite{diankov2010automated}.

\subsection{Solution approach}

We notice that the problem formulation in this section returns a constrained optimal 
base trajectory planning problem. 
Since time always marches forward and the mobile base is confined inside the 
admissible B-spacetime which guides the base from start to goal, the difficulty 
does not lie on exploring a region to find the path to goal but on constraints 
and optimization.
Therefore, we follow the discrete optimal planning approach \cite{lavalle2006planning} 
and propose a discrete optimal base trajectory planning algorithm based on dynamic 
programming.

\section{Optimal Base Trajectory Planning}

\subsection{Discretization}

We sample $N+1$ points along the desired end-effector trajectory with uniform 
time-step size $\Delta t = t_{end}/N$. 
Thus, the corresponding discrete base trajectory must also consist of $N+1$ points 
(time steps): $i = 0, ..., N$. 
At time step $i$, the time is $t^i = i \Delta t$ and the path parameter is $s^i = i/N$.

We discretize $\mathcal{V}_a$ by firstly specifying a set of velocity step 
sizes $\Delta v_x, \Delta v_y, \Delta \omega$ then finding combinations of 
integers $(k_x,k_y,k_{\varphi})$ such that the corresponding admissible spacetime velocities 
$\dot{\mathbf{x}} \coloneqq (1, k_x \Delta v_x, k_y \Delta v_y, k_{\varphi} \Delta \omega)$ 
satisfy the base velocity constraint in (\ref{eq:admiss_vel}). 
Then, the set of admissible controls is:
\begin{equation}
        \mathcal{U}_a = \{\mathbf{u} \coloneqq \dot{\mathbf{x}} \Delta t \;|\; 
        \dot{\mathbf{x}} \in \mathcal{V}_a\}
\end{equation}

We set the step sizes of the spatial coordinates at:
\begin{equation} \label{eq:spatial_discrete}
        \Delta x = \Delta v_x \Delta t, \quad
        \Delta y = \Delta v_y \Delta t, \quad
        \Delta \varphi = \Delta \omega \Delta t
\end{equation}
so that for any admissible controls, the mobile base moves from one grid point to another, 
i.e. 
\begin{equation}
        \mathbf{x}^{i+1} - \mathbf{x}^i = \dot{\mathbf{x}}^i \Delta t = \mathbf{u}^i
\end{equation}

\subsection{Kinematic Reachability Analysis}

\begin{figure}[tb]
        \centering
        \subfloat[Valid voxels cloud]{
                \includegraphics[width=0.47\linewidth]{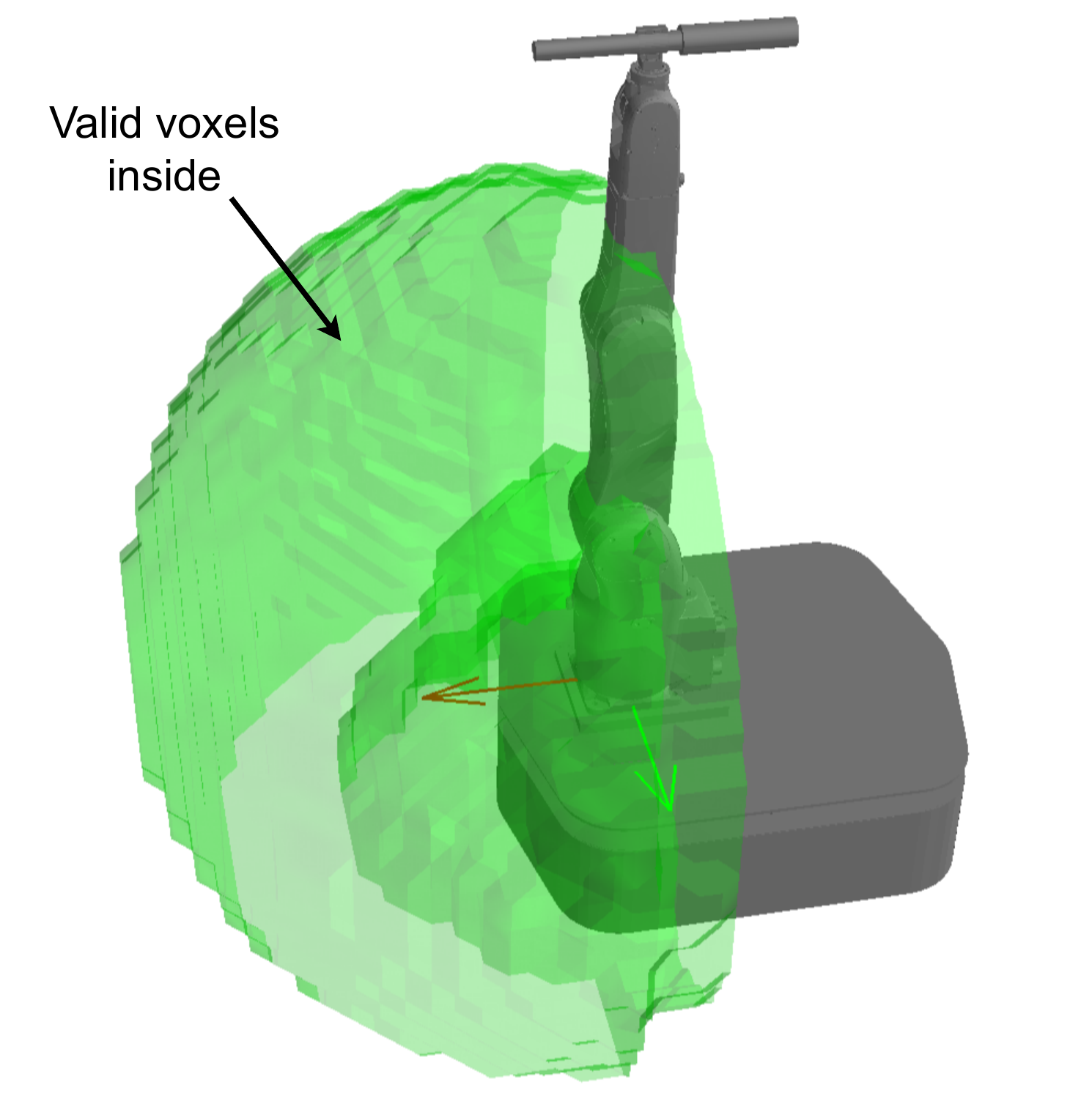}
                \label{fig:kr_raw}
        }
        \hfill
        \subfloat[Geometric reachable region]{
                \includegraphics[width=0.47\linewidth]{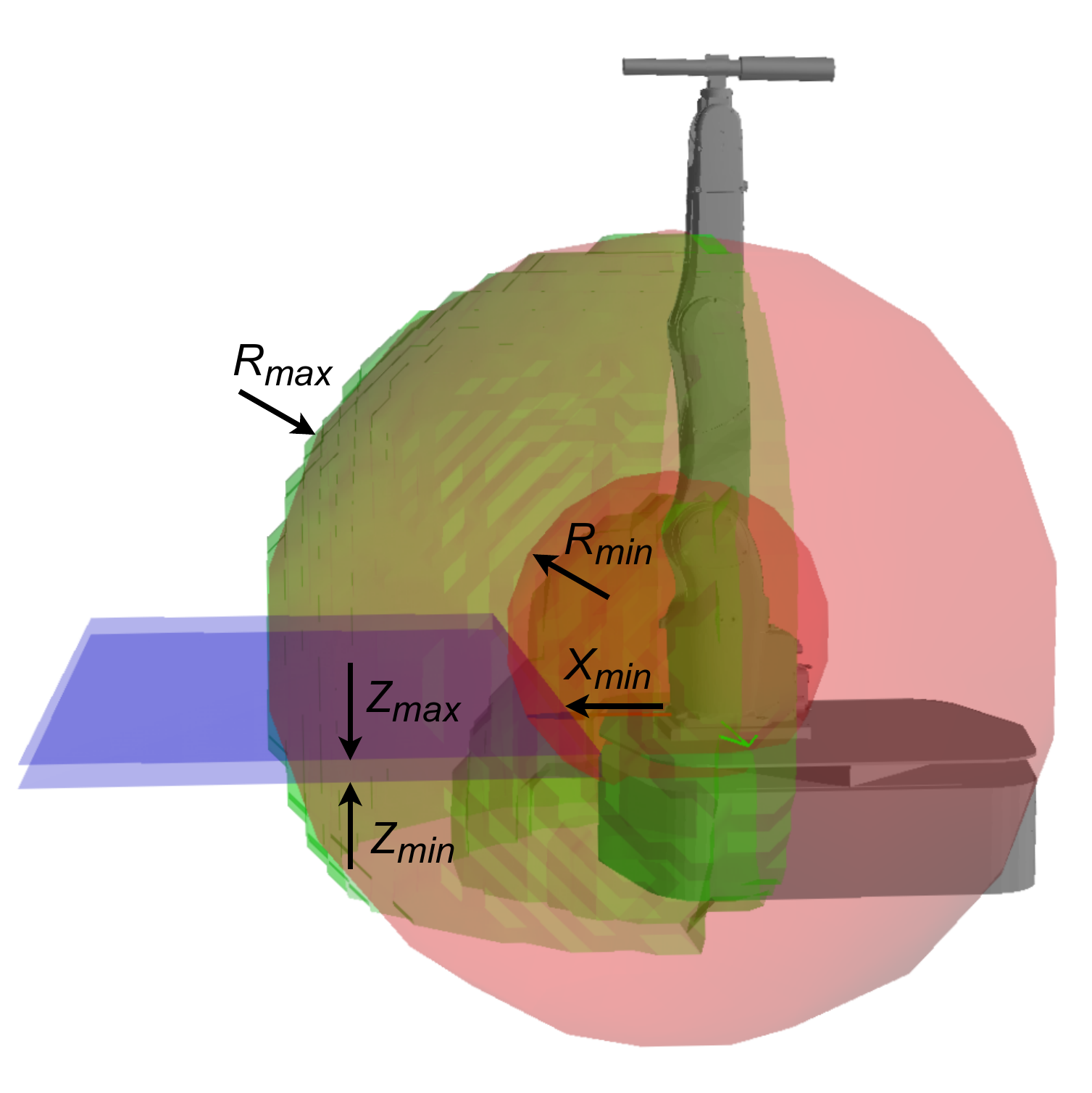}
                \label{fig:kr_analysed}
        }
        \caption{Visualization of Kinematic Reachability Analysis}
        \label{fig:kr}
\end{figure}

We implement the kinematic reachability analysis introduced in \cite{nguyen2023task} 
as follows. 
Firstly, we discretize the space relative to the manipulator into 3D voxels with 
size $\delta \times \delta \times \delta$.
Secondly, we specify a range of orientations based on the end-effector trajectory 
(geometric task constraints).
For example, in printing, the nozzle mainly points downwards, and we allow small 
deviation from this main direction. 
Next, we compute Inverse Kinematics and mark the valid voxels at which the 
end-effector can reach within the range of orientations determined previously.
At the end of this process, we obtain a valid voxel cloud (Fig. \ref{fig:kr_raw}) 
which can be stored and re-used for similar tasks with the same robot.

To use an obtained valid voxel cloud for a given task, we determine the 
\textit{geometric reachable region} according to each discretized point 
in the end-effector trajectory (a desired end-effector pose).
The geometric reachable region is obtained by slicing the valid voxels cloud using: 
2 horizontal planes bounding the height $h$ of the desired pose: $z = h \pm \delta/2$, 
1 vertical plane to keep a safe distance $X_{min}$ from the robot, 
and 2 spherical surfaces centred at the manipulator's second joint, 
with radii $R_{min}, R_{max}$ to maximize the volume of the in-between 
region containing only valid voxels (see Fig. \ref{fig:kr_analysed}).

\begin{figure}[tb]
        \centering
        \includegraphics[width=0.85\linewidth]{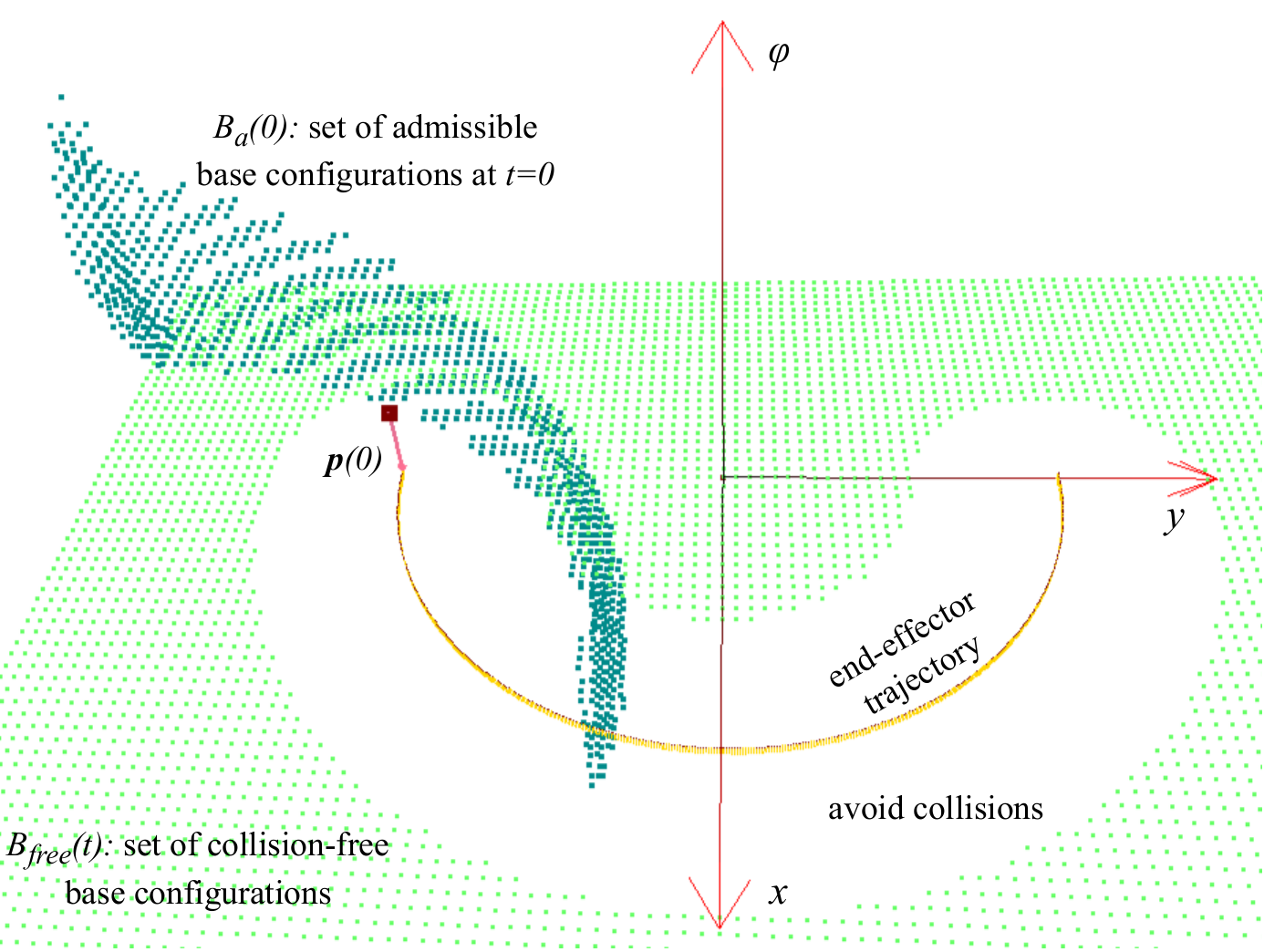}
        \caption{Discretized admissible B-space $\mathcal{B}_a(0)$ for end-effector pose $\mathbf{p}(0)$. 
        The vertical axis shows $\varphi \in (-\pi,\pi]$ for $\varphi \in \mathbb{S}^1$ 
        while time axis is not shown.}
        \label{fig:aBspace}
\end{figure}

Next, we obtain the discretized admissible B-space $\mathcal{B}_a^i$ 
(see Fig. \ref{fig:aBspace}) for each desired end-effector pose 
$\mathbf{p}(i \Delta t)$ as follows. 
Using the geometric reachable region parameters calculated for 
$\mathbf{p}(i \Delta t)$, we find all the collision-free base 
configurations where if the base is placed at, the desired end-effector pose 
is inside the geometric reachable region: 
$\mathbf{q}_b \in \mathcal{B}_a^i \subseteq \mathcal{B}_{free}(i \Delta t)$. 
Finally, the discretized admissible B-spacetime $\mathcal{X}_a$ is:
\begin{equation}
        \mathcal{X}_a = \bigcup_{i=0}^{N} \mathcal{X}_a^i, \quad
        \mathcal{X}_a^i = \{(i \Delta t, \mathbf{q}_b)\; |\; 
        \mathbf{q}_b \in \mathcal{B}_a^i\}
\end{equation}
This step can be computed in negligible time in our tests.

\subsection{Planning optimal mobile base trajectory}

The numerical problem is: find the minimum-cost trajectory in a multi-source 
($\mathbf{x}^0 \in \mathcal{X}_a^0$) multi-goal ($\mathbf{x}^N \in \mathcal{X}_a^N$) 
multi-stage ($i = 0, ..., N$) graph. 
The cost functional (\ref{eq:cost}) becomes: 
\begin{equation} \label{eq:total_cost}
        J = \sum_{i=0}^{N-1} l(\mathbf{u}^i)
        = \sum_{i=0}^{N-1} 
        \mathbf{u}^i \cdot \mathbf{I}_x \cdot \mathbf{u}^i \frac{1}{\Delta t}
\end{equation}

First, we introduce the following definitions which are inspired by similar 
concepts in control and planning \cite{pham2018new}.

\begin{definition}[One-step feasible set]
        \label{def:one_step_set}
        The \textit{one-step feasible set} $\mathcal{Q}(\mathcal{I})$ is a 
        set of admissible events $\mathbf{x} \in \mathcal{X}_a$ that can 
        reach at least one event $\bar{\mathbf{x}} \in \mathcal{I}$ using 
        one admissible control $\mathbf{u} \in \mathcal{U}_a$, that is
        $\bar{\mathbf{x}} = f(\mathbf{x}, \mathbf{u}) = \mathbf{x} + \mathbf{u}$ 
        (\textit{event transition equation})
        \begin{equation} \label{eq:one_step_set}
                \mathcal{Q}(\mathcal{I}) = \{ \mathbf{x} \;|\; 
                f(\mathbf{x}, \mathbf{u}) \in \mathcal{I}, \; 
                \mathbf{x} \in \mathcal{X}_a, \; 
                \mathbf{u} \in \mathcal{U}_a\}
        \end{equation}
\end{definition}

\begin{definition}[i-stage feasible set]
        \label{def:i_stage_feasible_set}
        The \textit{i-stage feasible set} $\mathcal{K}^i(\mathcal{I}^N)$ is 
        the set of admissible events that can reach at least one event in a 
        final set $\mathcal{I}^N$ after a sequence of $N-i$ admissible controls. 
        This set can be computed iteratively by 
        \begin{equation} \label{eq:control_set}
        \begin{aligned}
                &\mathcal{K}^N(\mathcal{I}^N) = \mathcal{I}^N \cap \mathcal{X}_a\\
                &\mathcal{K}^i(\mathcal{I}^N) = \mathcal{Q}(\mathcal{K}^{i+1}(\mathcal{I}^N))
        \end{aligned}
        \end{equation}
\end{definition}

In our case, the goal set is $\mathcal{I}_N = \mathcal{X}_a^N$ so 
(\ref{eq:control_set}) becomes
\begin{equation} \label{eq:connect}
\begin{aligned}
        &\mathcal{K}^N(\mathcal{X}_a^N) = \mathcal{X}_a^N \cap \mathcal{X}_a 
        = \mathcal{X}_a^N\\
        &\mathcal{K}^i(\mathcal{X}_a^N) = \mathcal{Q}(\mathcal{K}^{i+1}(\mathcal{X}_a^N)) 
        \subseteq \mathcal{X}_a^i
\end{aligned}
\end{equation}
which suggests a multistage graph where its stages are the same as the time 
steps of the trajectory $i = 0, 1, ..., N$, and every node in the graph lies 
on a grid point of the discretized admissible B-spacetime 
($\mathcal{K}^i \subseteq \mathcal{X}_a^i$).

Instead of constructing the graph before a subsequent graph search,
we propose running \textit{backward value iterations} as follows: 
while connecting the nodes using (\ref{eq:connect}), 
concurrently calculate the \textit{minimum cost-to-go} $G(\mathbf{x})$ 
\cite{lavalle2006planning} and store the \textit{optimal next-event} 
$\bar{\mathbf{x}}^*(\mathbf{x})$ (memoization) by:
\begin{equation} \label{eq:cost-to-go}
\begin{aligned}
        &G(\mathbf{x} \in \mathcal{K}^N) = 0 \\
        &G(\mathbf{x} \in \mathcal{K}^i) 
        = \min_{\bar{\mathbf{x}} \in \mathcal{K}^{i+1}} 
        \bigl\{ l(\bar{\mathbf{x}} - \mathbf{x}) + G(\bar{\mathbf{x}}) \;|\;
        \bar{\mathbf{x}} - \mathbf{x} \in \mathcal{U}_a \bigr\} \\
        &\bar{\mathbf{x}}^*(\mathbf{x} \in \mathcal{K}^i) 
        = \arg\min_{\bar{\mathbf{x}} \in \mathcal{K}^{i+1}} 
        \bigl\{ l(\bar{\mathbf{x}} - \mathbf{x}) + G(\bar{\mathbf{x}}) \;|\;
        \bar{\mathbf{x}} - \mathbf{x} \in \mathcal{U}_a \bigr\} \\
\end{aligned}
\end{equation}

The minimum cost $J_{min} = \min_{\mathbf{x}}\{G(\mathbf{x} \in \mathcal{K}^0)\}$ 
can be found as soon as the graph is fully connected, and the memoization can 
be used to recover the optimal trajectory. 
The whole procedure is summarized in Algorithm \ref{alg:trajopt} (MoboConTP).

For fast computation, we use coarse step sizes for Algorithm \ref{alg:trajopt} 
then apply linear interpolation to match with controller's rate, 
so the trajectory is piecewise $C^1$-continuous.

\subsection{Completeness and Optimality}

\begin{theorem}[Completeness] \label{theorem:completeness}
Algorithm \ref{alg:trajopt} only reports \texttt{Infeasible} 
when there is indeed no feasible trajectory.
\end{theorem}

\textit{Proof:} 
Since Algorithm \ref{alg:trajopt} only reports \texttt{Infeasible} when it 
runs into $\mathcal{K}^i = \emptyset$, we can prove by contradiction that: 
if there exists a feasible trajectory $\{\mathbf{x}^0, ..., \mathbf{x}^N\}$, 
then $\mathcal{K}^i$ contains $\mathbf{x}^i$ for all $i \in [0,N]$. 
Using backward induction:

\begin{itemize}
        \item Initialization: $\mathbf{x}^N \in \mathcal{K}^N$ by construction.
        \item Induction: Assume that $\mathbf{x}^{i+1} \in \mathcal{K}^{i+1}$; 
        Since $\mathbf{x}^i$ and $\mathbf{x}^{i+1}$ are in a feasible trajectory, 
        they satisfy $\mathbf{x}^{i+1} - \mathbf{x}^i \in \mathcal{U}_a$, 
        which means $\mathbf{x}^i \in \mathcal{Q}(\mathcal{K}^{i+1}) = \mathcal{K}^i$ 
        according to (\ref{eq:one_step_set}), (\ref{eq:connect}).
\end{itemize}
By induction, we get $\mathbf{x}^i \in \mathcal{K}^i \; \forall i \in [0,N]$, 
which means that Algorithm \ref{alg:trajopt} will not report \texttt{Infeasible} 
if a feasible trajectory exists.

\begin{theorem}[Optimality] \label{theorem:correct}
If Algorithm \ref{alg:trajopt} returns an output trajectory, 
then it is indeed the optimal trajectory.
\end{theorem}

\textit{Proof:} 
Firstly, if Algorithm \ref{alg:trajopt} returns a sequence 
$\{\mathbf{x}^0, ..., \mathbf{x}^N\}$, one can show by forward induction that 
this sequence is a feasible trajectory. 
Secondly, Algorithm \ref{alg:trajopt} tracks the minimum cost-to-go from each node 
to the goal stage in the same way as dynamic programming, 
so by the principle of optimality \cite{lavalle2006planning}, the returned 
sequence has the minimum cost-to-go from stage 0 to N, which means it is indeed 
the optimal trajectory.

\begin{algorithm}[t]
        \caption{MoboConTP - optimal base trajectory planning for 
        end-effector trajectory continuity} 
        \label{alg:trajopt}
        \small
        \DontPrintSemicolon
        \KwIn{Discretized admissible base configuration spacetime 
        $\mathcal{X}_a = \{\mathcal{X}_a^0, ..., \mathcal{X}_a^N\}$, 
        set of admissible controls $\mathcal{U}_a$}
        \KwOut{Optimal trajectory $\{\mathbf{x}^0,...,\mathbf{x}^N\}$}
        Initialization: 
        $\mathcal{K}^N \gets \mathcal{X}_a^N$ and 
        $G(\mathbf{x}) \gets 0 \; \forall \mathbf{x} \in \mathcal{K}^N$\;
        \tcc{Backward iterations}
        \For{$i \in [N-1, ..., 0]$}{
                $\mathcal{K}^i \gets \emptyset$\;
                \For{$\mathbf{x} \in \mathcal{X}_a^i$}{
                        $ValidNode \gets False$; $G(\mathbf{x}) \gets \infty$\; 
                        \For{$\bar{\mathbf{x}} \in \mathcal{K}^{i+1}$ 
                        such that $\bar{\mathbf{x}}-\mathbf{x} \in \mathcal{U}_a$}{
                                $ValidNode \gets True$\;
                                \If{$l(\bar{\mathbf{x}}-\mathbf{x}) + G(\bar{\mathbf{x}}) < G(\mathbf{x})$}{ 
                                        $G(\mathbf{x}) \gets l(\bar{\mathbf{x}}-\mathbf{x}) + G(\bar{\mathbf{x}})$\;
                                        $\bar{\mathbf{x}}^*(\mathbf{x}) \gets \bar{\mathbf{x}}$\;
                                }
                        }
                        \If{$ValidNode$}{
                                $\mathcal{K}^i.Insert(\mathbf{x})$
                        }
                }
                \If{$\mathcal{K}^i = \emptyset$}{
                        \Return{\texttt{Infeasible}}\;
                }
        }
        \tcc{Recover the optimal trajectory}
        $\mathbf{x}^0 \gets \arg\min_{\mathbf{x}} G(\mathbf{x} \in \mathcal{K}^0)$\;
        \For{$i \in [0, ..., N-1]$}{
                $\mathbf{x}^{i+1} \gets \bar{\mathbf{x}}^*(\mathbf{x}^i)$\;
        }
\end{algorithm}

\section{Evaluations}

Our mobile manipulator consists of a DENSO VS-087 6-DOF arm mounted on a 
Clearpath Ridgeback 3-DOF omnidirectional mobile base. 
The following experiment and simulations illustrate our method in mobile printing 
at different difficulty levels, all were performed using CPU AMD 5900HX with 
16GB RAM, running ROS Melodic in Ubuntu 18.04. 
Simulations were performed in OpenRAVE environment \cite{diankov2010automated}. 
A hardware demo and a simulation example are shown in the accompanying video 
(https://youtu.be/yyBv3xGClnk).

\subsection{Method visualization: Mobile 1D printing}

\begin{figure*}[tb]
        \centering
        \subfloat[1D mobile base ($x=\varphi=0$)]{
                \includegraphics[width=0.32\linewidth]{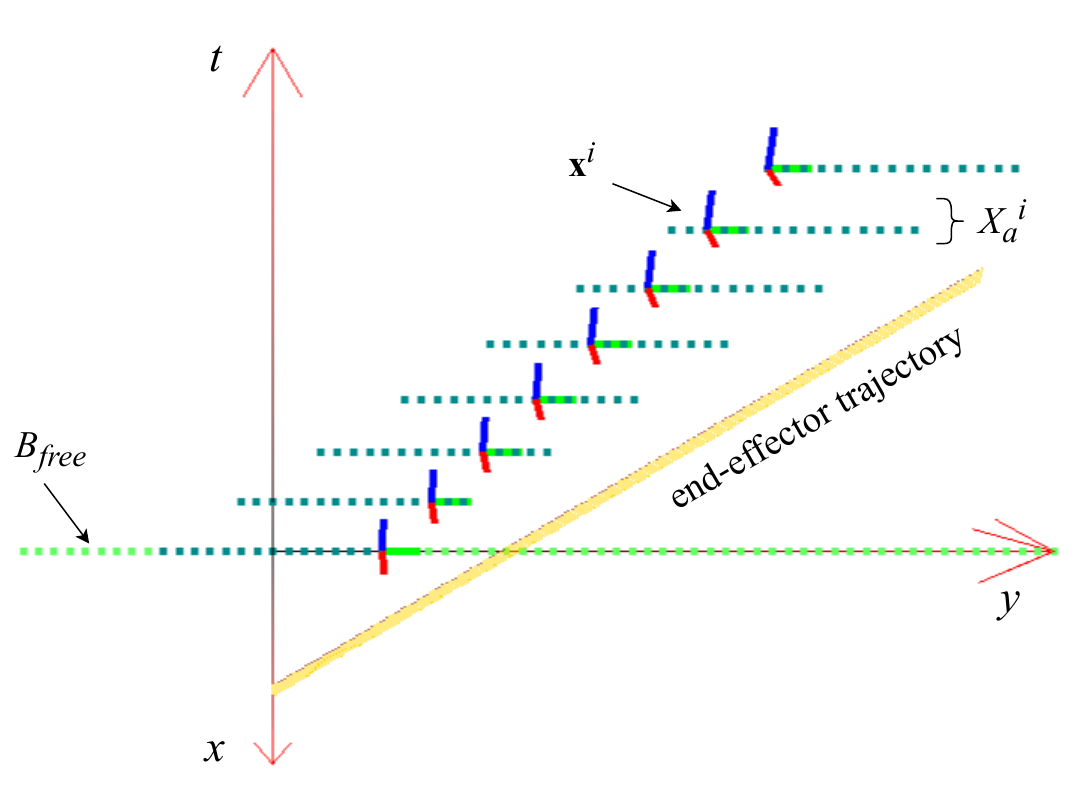}
                \label{fig:1Dprinting_1Dbase}
        }
        \hfill
        \subfloat[2D mobile base ($\varphi=0$)]{
                \includegraphics[width=0.24\linewidth]{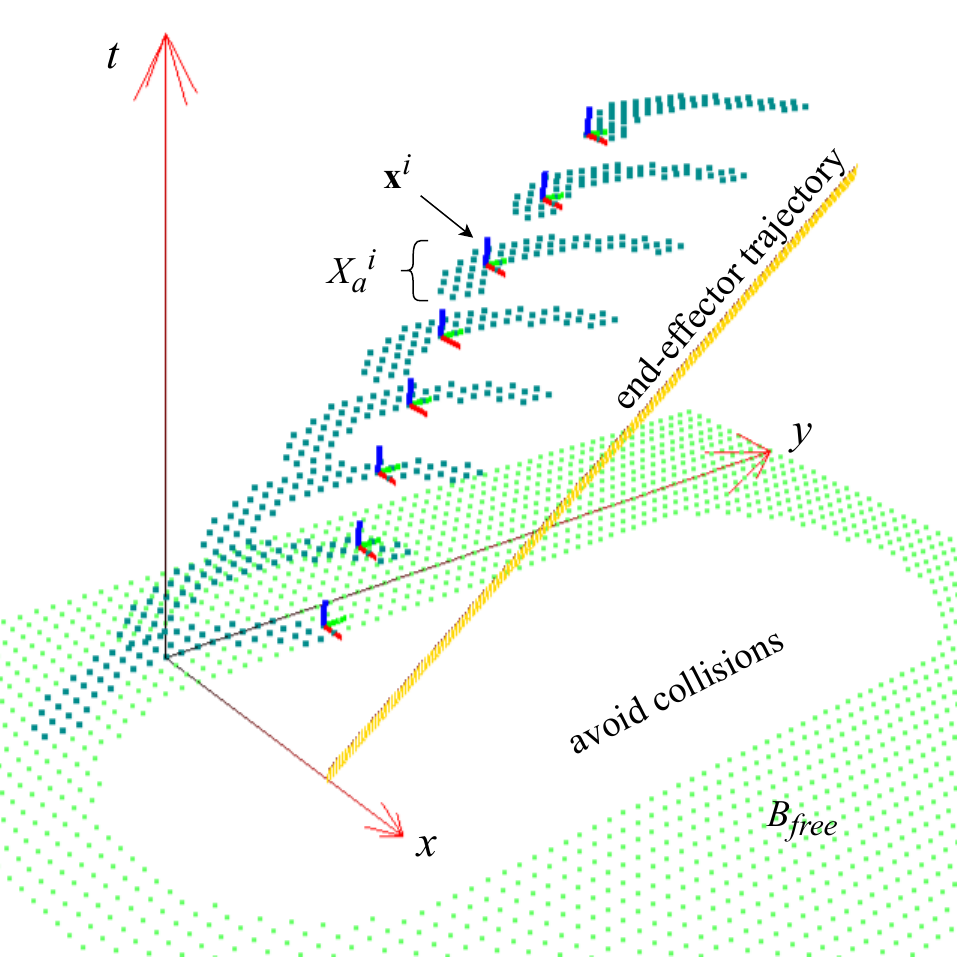}
        }
        \hfill
        \subfloat[Omnidirectional 3D planar mobile base]{
                \includegraphics[width=0.36\linewidth]{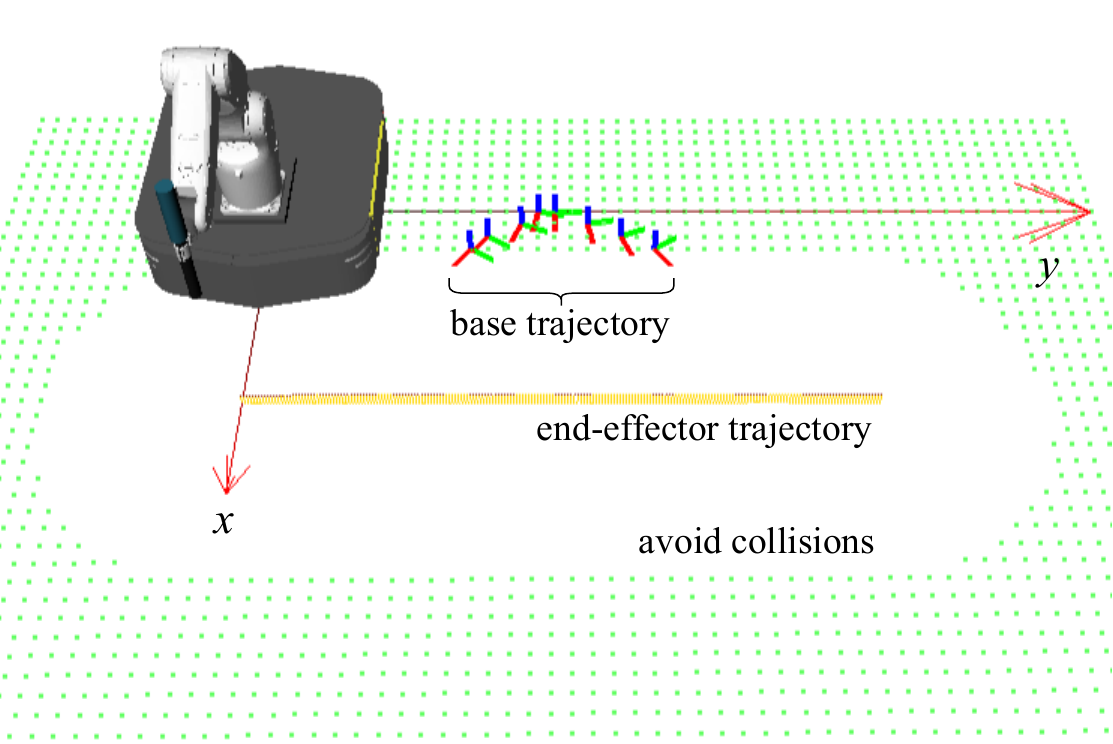}
                \label{fig:1Dprinting_3Dbase}
        }
        \caption{Mobile 1D printing a $2.1m$-long line with different base's DOFs.
        (a) and (b) are seen in B-spacetime, (c) is seen in task space.}
        \label{fig:1Dprinting}
\end{figure*}

Fig. \ref{fig:1Dprinting} shows the optimal base trajectory solutions 
for 3 cases of mobile base in 1D printing a horizontal line.
Since the B-spacetime for 3D base is 4-dimensional ($t,x,y,\varphi$), Fig. 
\ref{fig:1Dprinting_3Dbase} can only show the solution trajectory without time axis.

\subsection{Algorithm benchmarking: Mobile 3D printing}

Next, we benchmark MoboConTP algorithm and compare it with a baseline (Dijkstra's 
algorithm) in a 3D printing task as shown in Fig. \ref{fig:3Dprinting_NTU}: 
an NTU shape with the total printing path length of $d=112.9m$ (10 layers).
Table \ref{tab:comparison} compares the computation time of two methods (both using 
$\Delta t=2.5s$, $\Delta v_x = \Delta v_y = 5cm/s$, $\Delta \omega = \pi/30\,rad/s$). 
MoboConTP is faster by running optimization and graph construction concurrently. 

\begin{table}[b]
        \caption{Planning time comparison (in NTU test, Fig. \ref{fig:3Dprinting_NTU})}
        \label{tab:comparison}
        \begin{center}
        \begin{tabular}{|c|c c c c|}
        \hline
        Planning time for: & 8 layers & 10 layers & 12 layers & 14 layers\\
        \hline
        MoboConTP & 89.9s & 108.4s & 129.7s & 155.1s\\
        Baseline (Dijkstra) & 120.3s & 155.2s & 184.4s & 515.0s\\
        \hline
        \end{tabular}
        \end{center}
\end{table}

\subsection{Comparison: Mobile 3D printing}

\begin{figure}[tb]
        \centering
        \includegraphics[width=0.45\linewidth]{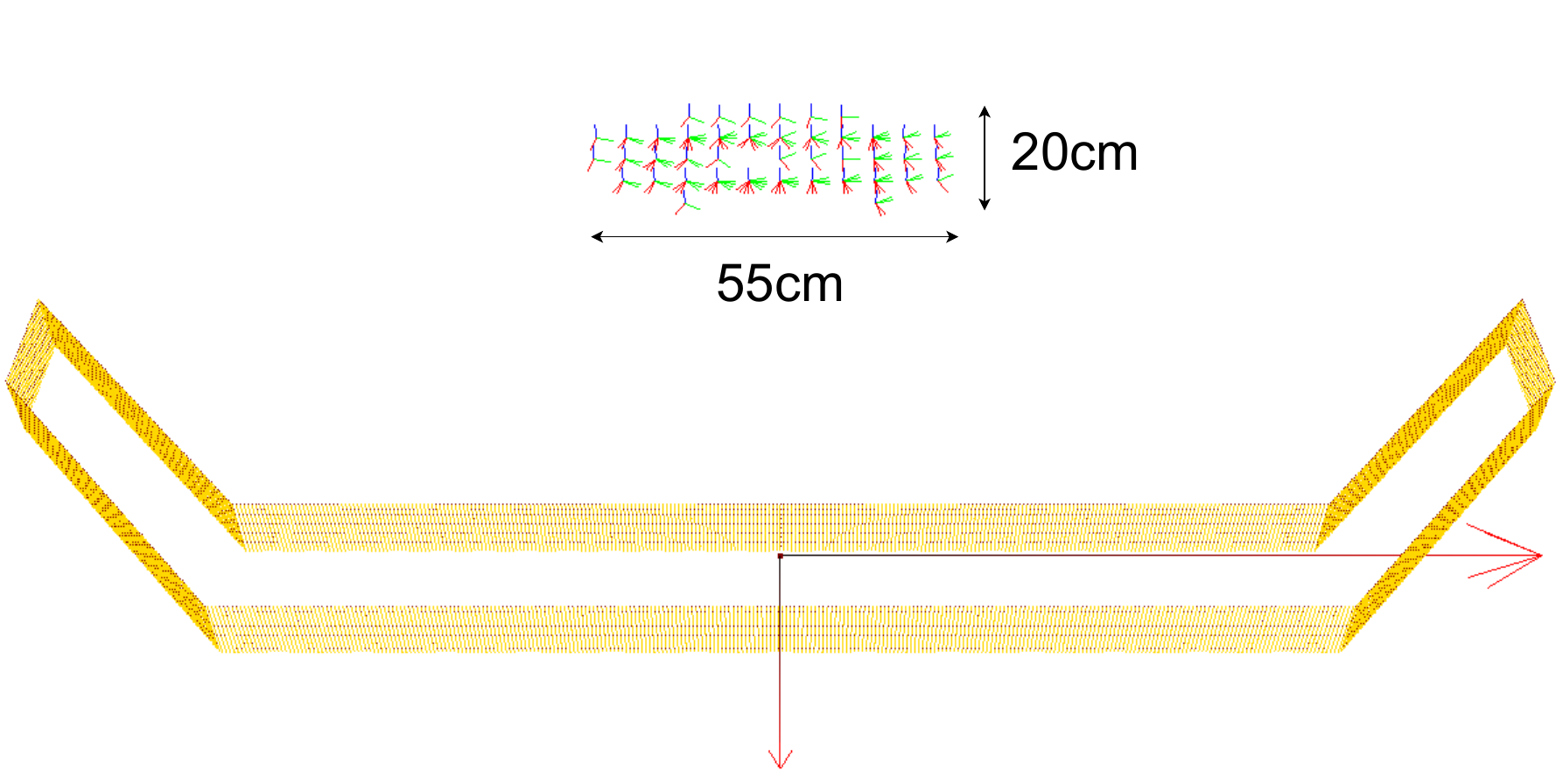}
        \caption{Optimal base trajectory for mobile 3D printing task in \cite{tiryaki2019printing}}
        \label{fig:3Dprinting_compare}
\end{figure}

We would like to compare our method with manual planning in \cite{tiryaki2019printing}. 
Fig. \ref{fig:3Dprinting_compare} shows our solution where the mobile base 
moves in a region $55\times20cm$ instead of $80\times10cm$, with an average 
speed $2cm/s$ which is lower than $3.5cm/s$ in \cite{tiryaki2019printing}.

\subsection{Hardware demo: Mobile 3D printing}

\begin{figure}[tb]
        \centering
        \includegraphics[width=0.85\linewidth]{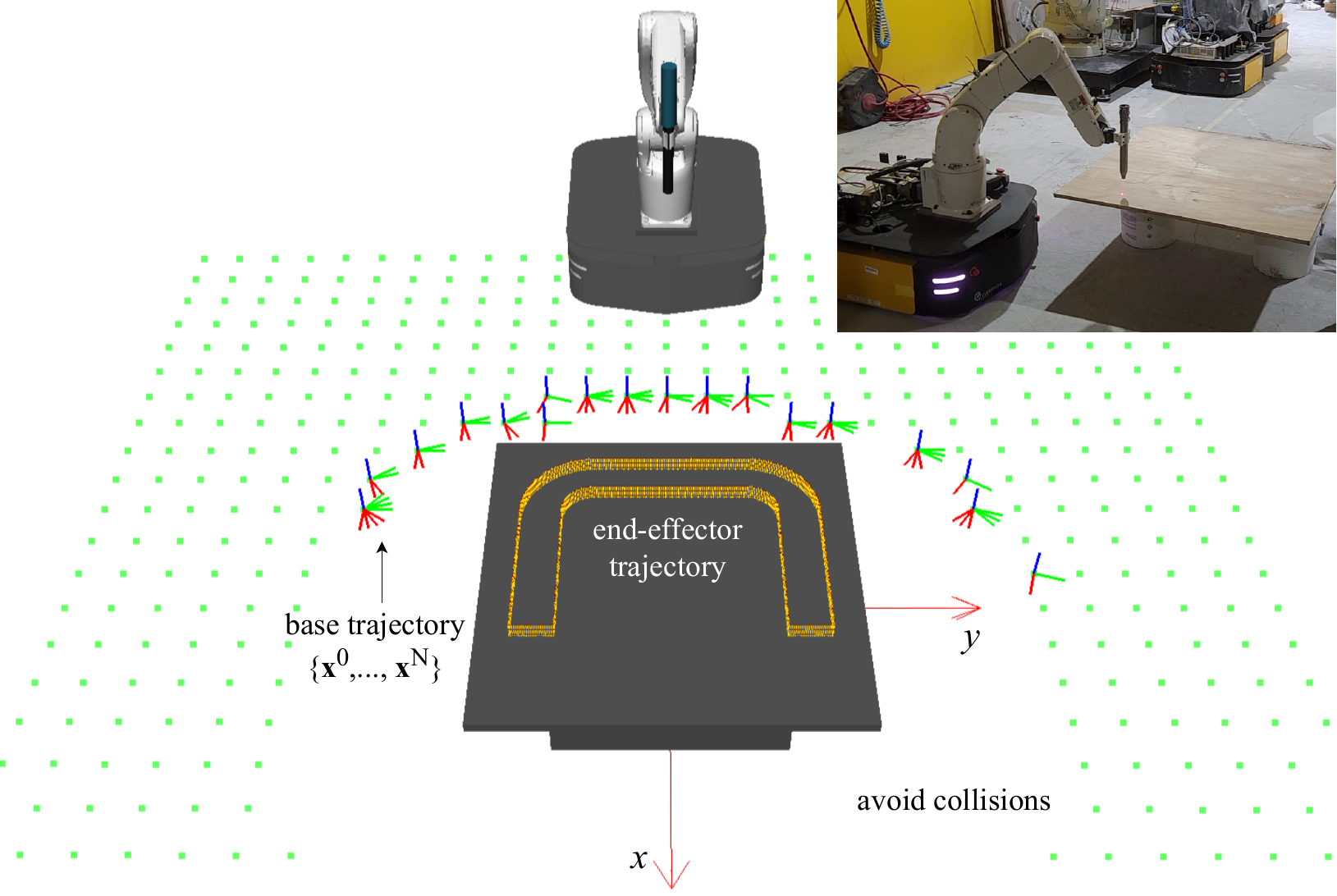}
        \caption{Hardware mobile 3D printing demo: U shape ($0.9 \times 0.675 \times 0.05m$, 
        5 layers, total printing path: $19.85m$) with constant nozzle speed of $10cm/s$.}
        \label{fig:3Dprinting_U}
\end{figure}

Fig. \ref{fig:3Dprinting_U} shows a hardware demo of a 3D printing task 
(5 layers of U shape) with total printing path length $d=19.85m$.
Our algorithm planned an optimal mobile base trajectory in $3.8s$ (using 
$\Delta t = 3s$, $\Delta v_x = \Delta v_y = 5cm/s$, $\Delta \omega = \pi/30\,rad/s$).

\subsection{Complexity analysis}

\subsubsection{Complexity with respect to discretization step sizes}
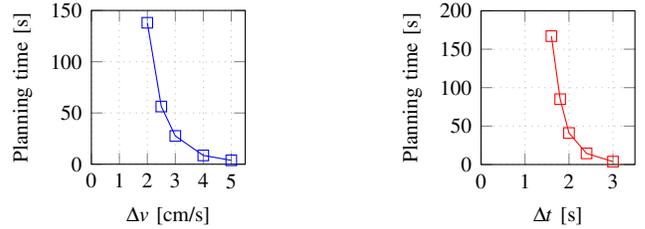
\begin{figure}[tb]
\centering
\subfloat[Planning time for U shape (5 layers), using $\Delta t = 3s$]{
        \begin{tikzpicture}
        \begin{axis}[
                xlabel={$\Delta v$ [cm/s]},
                ylabel={Planning time [s]},
                xmin=0, xmax=5.5,
                ymin=0, ymax=150,
                xtick={0,1,2,3,4,5},
                ytick={0,50,100,150},
                grid style=dashed,
                width=0.42\linewidth,
                height=0.42\linewidth,
                font=\footnotesize,
                xmajorgrids=true,
                ymajorgrids=true,
                grid style=dotted
        ]
        \addplot[
                color=blue,
                mark=square,
                ]
                coordinates {
                (2,138)(2.5,56.3)(3,27.6)(4,8.6)(5,3.8)
                };
        \end{axis}
        \end{tikzpicture}
}
\hfill
\subfloat[Planning time for U shape (5 layers), using $\Delta v = 5cm/s$]{
        \begin{tikzpicture}
        \begin{axis}[
                xlabel={$\Delta t$ [s]},
                ylabel={Planning time [s]},
                xmin=0, xmax=3.5,
                ymin=0, ymax=200,
                xtick={0,1,2,3},
                ytick={0,50,100,150,200},
                grid style=dashed,
                width=0.42\linewidth,
                height=0.42\linewidth,
                font=\footnotesize,
                xmajorgrids=true,
                ymajorgrids=true,
                grid style=dotted
        ]
        \addplot[
                color=red,
                mark=square,
                ]
                coordinates {
                (1.6,167)(1.8,85)(2,41)(2.4,14.5)(3,3.8)
                };
        \end{axis}
        \end{tikzpicture}
}
\caption{Complexity of Algorithm \ref{alg:trajopt} with respect to discretization step sizes}
\label{fig:benchmark}
\end{figure}

Number of stages is $N \propto 1/ \Delta t$; number of nodes per stage is 
$|\mathcal{X}_a^i| \propto (1/ \Delta x \Delta y \Delta \varphi)
\propto (1/ \Delta t^3 \Delta v_x \Delta v_y \Delta \omega)$.
Based on Algorithm \ref{alg:trajopt}, we expect its complexity to be no higher than 
$O \left( (1/ \Delta t)^7 (1/ \Delta v_x \Delta v_y \Delta \omega)^2 \right)$.
Fig. \ref{fig:benchmark} shows in our test case that, as 
$\Delta v \coloneqq \Delta v_x = \Delta v_y$ and $\Delta t$ decrease, 
the computation time increases polynomially with order 
$O((1/\Delta v)^{3.9})$ and $O((1/\Delta t)^{6.0})$ respectively 
(based on log-log analysis).

\subsubsection{Complexity with respect to task length}
Table \ref{tab:comparison} shows that the computation time of Algorithm 
\ref{alg:trajopt} increases linearly, since $N \propto d$. 
This is crucial for large-scale applications.

\section{Conclusion}

In this paper, we have proposed a method for solving the mobile manipulator 
motion planning problem under end-effector trajectory continuity constraint, 
subject to other constraints and base trajectory optimization. 
We have also presented a complete and optimal algorithm to plan optimal 
base trajectories, which was evaluated in several mobile printing tasks. 
Our future directions include: integrating planning with control for actual mobile 
concrete printing and accuracy evaluations, and addressing the following limitations:
\begin{itemize}
        \item The joint kinematic constraints were not considered. 
        \item The base acceleration constraints were not considered.
        \item During manipulator trajectory planning, although continuous joint 
        trajectories were obtained in our tests, this may not be guaranteed for 
        other tasks or robots.
        \item The base trajectory planning algorithm can be improved using other 
        trajectory optimization methods.
        \item Infeasible tasks can be tackled by a multi-robot team.
\end{itemize}

\section*{Acknowledgment}

This research was supported by the National Research Foundation, Prime Minister's 
Office, Singapore under its Medium-Sized Centre funding scheme, 
CES\textunderscore SDC Pte Ltd, and Sembcorp Architects \& Engineers Pte Ltd.

\addtolength{\textheight}{-7.3cm}

\bibliographystyle{IEEEtran}
\bibliography{IEEEabrv, references}

\end{document}